\documentclass[12pt,a4paper]{article}

\usepackage{natbib}
\usepackage[utf8]{inputenc}
\usepackage[T1]{fontenc}
\usepackage[brazil,english]{babel}
\usepackage{geometry}
\usepackage{setspace}
\usepackage{graphicx}
\usepackage{float}
\usepackage{amsmath,amssymb}
\usepackage{booktabs}
\usepackage{accessibility}
\usepackage{longtable}
\usepackage{array}
\usepackage{hyperref}
\usepackage{xcolor}
\usepackage{listings}
\usepackage{caption}
\usepackage{tikz}
\usepackage{placeins}

\lstdefinestyle{plain}{
  basicstyle=\ttfamily\small,
  numbers=left,
  numberstyle=\tiny,
  stepnumber=1,
  numbersep=6pt,
  frame=single,
  breaklines=true,
  showstringspaces=false,
  tabsize=2
}

\lstdefinestyle{json}{
  style=plain,
  backgroundcolor=\color{gray!10}
}

\lstdefinestyle{python}{
  style=plain,
  language=Python,
  backgroundcolor=\color{gray!10}
}

\lstdefinestyle{policy}{
  style=plain,
  backgroundcolor=\color{gray!10}
}

\title{
\textbf{Liquid Interfaces: A Dynamic Ontology for the Interoperability of Autonomous Systems}\\
}

\author{
Dhiogo José Correa de Sá\thanks{\href{mailto:dhiogo.correa@draiven.io}{dhiogo.correa@draiven.io}}
\and
Carlos Pereira Lopes Filho\thanks{\href{mailto:carlos.lopes@draiven.io}{carlos.lopes@draiven.io}}
\and
Carlos Eduardo Schmiedel\thanks{\href{mailto:carlos.schmiedel@draiven.io}{carlos.schmiedel@draiven.io}}
}

\date{2026}

\begin{document}
\maketitle

\begin{abstract}
Contemporary software architectures struggle to support autonomous agents whose reasoning is adaptive, probabilistic, and context-dependent, while system integration remains dominated by static interfaces and deterministic contracts. This paper introduces Liquid Interfaces, a coordination paradigm in which interfaces are not persistent technical artifacts, but ephemeral relational events that emerge through intention articulation and semantic negotiation at runtime.

We formalize this model and present the Liquid Interface Protocol (LIP), which governs intention-driven interaction, negotiated execution, and enforced ephemerality under semantic uncertainty. We further discuss the governance implications of this approach and describe a reference architecture that demonstrates practical feasibility. Liquid Interfaces provide a principled foundation for adaptive coordination in agent-based systems.
\end{abstract}


\section{Introduction}

The history of system integration is fundamentally a chronicle of attempts to impose order upon informational chaos. From nascent Remote Procedure Call (RPC) protocols \cite{birrell1984implementing} to the ubiquity of REST \cite{fielding2000} and GraphQL \cite{hartig2018semantics}, the overarching objective has been standardization. The foundational premise of traditional Software Engineering posits that security and efficiency are contingent upon predictability: System A must determine, with byte-level precision, what System B anticipates receiving. This paradigm, codified by Fielding’s architectural constraints \cite{fielding2000}, has succeeded for decades, enabling the construction of robust, complex systems through explicit and immutable contracts.

However, the integration of Large Language Models (LLMs) and Autonomous Agents into enterprise workflows has precipitated a fundamental ontological dissonance \cite{zhang2025selforganizing}. On one spectrum lies the liquid agent, which operates probabilistically, interprets natural language, adapts to ambiguous contexts, and pursues high-level objectives (e.g., ``reduce logistics costs by 15$\%$ while maintaining delivery quality'') \cite{xi2023agents}. On the other lies the ``solid API,'' which operates deterministically, demands rigid syntax, and executes discrete functions (e.g., \texttt{POST /orders/v1/create} with a predefined JSON schema). This friction is not merely technical; it is ontological. As highlighted by recent research on tool-augmented language models, the capability of agents to interact with external tools is often constrained by the rigid structural requirements of existing interfaces \cite{schick2024toolformer}.

Consider a practical scenario: an AI agent within a logistics firm must respond to a storm that has closed a port. The agent perceives the disruption, yet the legacy routing system's API merely returns a static error: ``Route Unavailable.'' The agent cannot negotiate alternatives; it lacks the protocol to converse with disparate systems to synthesize a solution. It remains trapped in a deterministic impasse. Consequently, a human operator must intervene, manually querying siloed systems, negotiating with suppliers, and updating multiple platforms. What should be an autonomous decision devolves into a costly, manual process.

The cost of translation between the agent's fluid intent'' and the API's rigid form'' has emerged as the primary bottleneck for intelligent automation. The maintenance of connectors, the synchronization of documentation (e.g., Swagger/OpenAPI), and the management of breaking changes are symptomatic of a paradigm that has reached its elastic limit. As noted in the context of machine learning systems, these integration glues and schema dependencies contribute significantly to hidden technical debt, compounding exponentially with each new system addition \cite{sculley2015hidden}.

To transcend these limitations, we propose the \textbf{Liquid Interface}: a framework that eschews static form as a prerequisite for interaction, privileging flow and intent. Rather than rigid systems connected via static contracts, we propose a paradigm wherein interfaces emerge dynamically, negotiated at runtime between agents that may possess no prior history of interaction.

A Liquid Interface is not an endpoint, a contract, or a schema. It is a relational event that manifests when two or more agents require collaboration. It arises from necessity, stabilizes only for the required duration, and dissolves once its purpose is fulfilled, leaving zero technical debt. In this model, the logistics agent could express its intent: ``I need to deliver this container within a 200km radius of the closed port, optimizing for cost, time, and delivery quality.'' The system would autonomously discover which agents (warehouses, carriers, suppliers) could assist, negotiate terms, and execute the solution—all without human intervention or pre-written integration code.

This paper intentionally focuses on the conceptual, formal, and architectural foundations of Liquid Interfaces. Concrete implementations, performance trade-offs, and empirical evaluations are treated as separate concerns and are the subject of ongoing and future work.
\section{Theoretical Foundation and Related Work}

Liquid Interfaces are proposed as an architectural response to the increasing heterogeneity, uncertainty, and volatility found in modern distributed computational environments. While existing paradigms provide robust abstractions for integrating well-defined services with stable contracts, they struggle in settings where participants are unknown at design-time, capabilities evolve dynamically, and meaning must be negotiated during execution. In such settings, \emph{static} agreement mechanisms become sources of friction rather than enablers of coordination.

\subsection{Architectural Motivation}

Most contemporary distributed systems assume that interacting components can establish a stable representational contract prior to execution. Microservices, Service Meshes, and API Gateways operate under this assumption by enforcing static schemas, predefined endpoints, and lifecycle policies that support syntactic interoperability. Semantic Web technologies extend this premise into the semantic domain by attempting to align vocabularies through global ontologies and shared taxonomies. Agent Communication Languages (ACLs) further assume that interacting agents share a common logical framework and can interpret performative acts within that logic.

These architectures are effective when the system is closed, the interfaces are known in advance, and the domain is sufficiently stable. However, their design assumptions introduce friction in open-ended computational ecosystems in which (i) novel services may appear without prior registration, (ii) capabilities may vary over time due to learning or context, and (iii) goals may be situational rather than predetermined. In these environments, static schemas and global ontologies impose upfront alignment costs that scale poorly and reduce the flexibility of the agentic ecosystem.

\subsection{Limits of LLM-Based Agentic Systems}

Recent research demonstrates that Large Language Models (LLMs) can autonomously plan, invoke tools, and orchestrate interactions across heterogeneous systems, effectively acting as general-purpose policy engines. While this represents a major advance in agentic architectures, scaling LLMs does not eliminate several intrinsic limitations. Recent theoretical work formalizes five such constraints~\cite{mohsin2025fundamental}, namely: (i) inevitable hallucination due to inherent approximation error, (ii) context compression limitations that degrade reasoning over long inputs, (iii) brittle reasoning emerging from likelihood-based training, (iv) retrieval fragility caused by semantic mismatch between evidence and generation, and (v) multimodal misalignment that undermines robust cross-signal grounding. Empirical studies corroborate several of these phenomena: LLMs exhibit irreducible uncertainty even when confident~\cite{kadavath2022language}, and their ability to exploit long contexts deteriorates as distance from the query increases~\cite{liu2023lost}.

These limitations have two consequences for open agentic ecosystems. First, they imply that agentic systems cannot rely on deterministic correctness or complete semantic grounding, even when augmented with planning or retrieval mechanisms. Second, they indicate that LLMs can negotiate actions, but lack mechanisms for negotiating interfaces or meaning under ambiguity, especially when semantic context is asymmetrically distributed among participants. In other words, while LLMs can decide what to do, they struggle to decide how to talk about what is being done when no prior agreement exists.

\subsection{Proposal: Liquid Interfaces}

The Liquid Interface Protocol (LIP) addresses this gap by shifting interface formation from \emph{static contract definition} to \emph{dynamic semantic negotiation}. Instead of requiring global agreement before execution, LIP enables autonomous agents to (i) discover capabilities at runtime, (ii) negotiate intent without shared ontologies, (iii) align meaning through iterative clarification, and (iv) dissolve interfaces once goals are achieved. Under this perspective, interfaces are not stable artifacts but emergent relational constructs optimized for adaptability during execution.

This reframing is relevant for ecosystems where the cost of prior semantic alignment is prohibitive or where the domain shifts faster than global ontologies can be updated. Liquid Interfaces do not replace existing contract-based architectures, but complement them by enabling a spectrum of coordination modes ranging from fully-static to fully-dynamic. Static contracts remain beneficial when stability is desired; Liquid Interfaces become advantageous when flexibility is required.

\subsection{Related Work and Differentiation}

The proposed architecture situates itself at the intersection of several established domains. While it draws upon foundational concepts from distributed systems and semantic computing, it diverges fundamentally in its handling of uncertainty and interface adaptability. Table \ref{tab:comparison} synthesizes the primary distinctions between traditional paradigms and the proposed Liquid Interfaces (LIP), delineating the specific contributions of this work.

\begin{table}[H]
\centering
\caption{Comparison with Traditional Approaches and Research Contributions}
\label{tab:comparison}
\small 
\begin{tabular}{p{3.0cm} p{5.5cm} p{5.5cm}}
\toprule
\textbf{Research Area} & \textbf{Traditional Approach} & \textbf{Liquid Interfaces Contribution} \\
\midrule
Semantic Web \& Ontologies &
Focuses on shared global vocabularies (RDF, OWL) and static ontology mapping to achieve a universal truth \cite{BernersLee2001}. &
Abandons the requirement for a pre-agreed global lexicon, focusing instead on run-time meaning negotiation and the generation of ephemeral, context-dependent ontologies. \\
\addlinespace
Microservices \& Service Mesh &
Utilizes orchestrators (e.g., Kubernetes) and meshes (e.g., Istio) to manage syntactic communication between services with well-defined APIs \cite{Posta2023}. &
Operates at a higher level of abstraction, orchestrating \textit{intentions} rather than services. The interface is not a fixed endpoint, but an emergent relational pact. \\
\addlinespace
Agent Communication Languages (ACL) &
Defines rigid performative message standards (e.g., FIPA-ACL) for rational agents to interact based on formal logic \cite{FIPA2002}. &
Replaces formal logic with probabilistic language models and computational hermeneutics, enabling interaction between non-deterministic agents without strict protocol compliance. \\
\addlinespace
API Management \& Gateways &
Centralizes access control and routing based on static contracts (OpenAPI/Swagger). &
The Semantic Service Bus acts not as a contract gatekeeper, but as an ontological mediator facilitating dynamic capability discovery and translation. \\
\addlinespace
Smart Contracts (Blockchain) &
Automates the execution of immutable, deterministic agreements in a decentralized environment \cite{Buterin2014}. &
Proposes ephemeral, adaptive contracts that dissolve post-execution. The focus is on agility and negotiation rather than immutability and trustless verification. \\
\addlinespace
Model Context Protocol (MCP) &
Standardizes the exposure of local tools and context to LLMs, answering the question ``what is available?'' \cite{MCP2024}. &
Complementary to MCP, it answers ``what should happen?'' by orchestrating intentions across multiple agents. MCP provides local technical contracts; LIP governs global coordination. \\
\bottomrule
\end{tabular}
\end{table}

Semantic Web aspires to global semantic convergence, whereas LIP enables local reconciliation of meaning under partial knowledge. Service Meshes optimize syntactic reliability between predefined components; LIP supports semantic discovery and negotiation among heterogeneous autonomous agents. Classical ACLs assume shared logical formalisms; LIP tolerates heterogeneous reasoning paradigms based on probabilistic interpretation and contextual inference. Smart Contracts emphasize immutability and determinism; Liquid Interfaces emphasize ephemerality and negotiated action. Compared to the Model Context Protocol (MCP), which answers ``what is available?'', LIP answers ``what should happen?'' by orchestrating intent across multiple agents.

These differences suggest that Liquid Interfaces occupy a complementary design space situated between deterministic protocol engineering and open-ended agentic negotiation.

\subsection{Interpretive Lens}

Although the proposal is technical in nature, it can also be situated within broader interpretive frameworks that clarify the motivations behind its emergence and help articulate its departure from traditional architectural paradigms. Bauman's notion of \emph{liquid modernity} describes a socio-historical shift in which stable, durable, and institutionally enforced structures give way to arrangements characterized by fluidity, adaptability, and temporary association~\cite{Bauman2000}. In Bauman's analysis, the ``solid'' phase of modernity was defined by an emphasis on planning, standardization, and long-term institutional commitments, whereas the ``liquid'' phase privileges flexibility, reconfiguration, and short-lived alignments that dissolve once they cease to serve immediate purposes.

When interpreted computationally, this lens highlights the growing misalignment between rigid digital infrastructures---which operate through static contracts, global ontologies, and pre-negotiated schemas---and contemporary computational ecosystems that are increasingly open, heterogeneous, and fast-moving. Under this framing, architectural rigidity becomes a liability: just as Bauman observes that solid structures struggle to persist under rapidly changing conditions, static computational protocols incur high coordination costs as system volatility increases. The consequence is an impedance mismatch between the demands of the environment and the affordances of the underlying architecture.

Gadamer's concept of \emph{fusion of horizons} offers a complementary perspective on semantic negotiation. Rather than assuming that meaning is fixed or universally shared, Gadamer posits that understanding emerges through dialogue between heterogeneous actors, each bounded by its own horizon of assumptions, intentions, and capabilities~\cite{Gadamer1960}. Liquid Interfaces adopt a similar stance by allowing interacting agents to converge on temporary semantic alignments without requiring global ontological consensus.

Actor--Network Theory (ANT)~\cite{Latour2005} further extends this view by collapsing the classical subject--object distinction and treating both human and non-human entities as \emph{actants} embedded within sociotechnical assemblages whose stability is not given, but continuously negotiated through association. Under this interpretation, the interface becomes a momentary pact that sustains coordination, dissolving once the underlying relations no longer hold.

These conceptual lenses are not prerequisites for implementation, nor do they constitute the core technical contribution of Liquid Interfaces. Rather, they provide a vocabulary that makes explicit why architectures oriented around runtime negotiation, semantic adaptability, and ephemeral coordination become desirable as computational ecosystems transition from closed, highly structured environments to open, heterogeneous, and rapidly shifting ones.

\section{Formal Definition: From Object to Event}

The transition from the solid to the liquid paradigm represents an ontological shift from static structural artifacts to dynamic temporal events. This section provides the mathematical formalization of this transition, defining the Liquid Interface not as a persistent boundary, but as a transient solution to a negotiation problem.

\subsection{The Solid Paradigm: Static Contractual Isomorphism}

In the classical paradigm, an interface $I_{\text{solid}}$ is defined as a static contract enforcing strict structural compliance. Let $\mathcal{S}$ be the universal set of possible schemas and $\Sigma \subset \mathcal{S}$ be a specific, pre-defined schema. We define the solid interface as a tuple:

\begin{equation}
    I_{\text{solid}} = (\Sigma, \mathcal{F}, \lambda) \quad
\end{equation}

Where:
\begin{itemize}
    \item $\Sigma$ represents the fixed data structure (ontology/schema).
    \item $\mathcal{F}$ is the set of permissible operations (endpoints/methods).
    \item $\lambda: X \to \{0, 1\}$ is a validation function (the contract enforcement).
\end{itemize}

The interaction is considered valid if and only if the input vector $\vec{x}$ satisfies the strict compliance condition defined by Meyer's \textit{Design by Contract} principles \cite{Meyer1992}:

\begin{equation}
    forall \bar{x} \in X_{\text{input}}, \quad \text{Valid}(\bar{x}) \iff \bar{x} \in \text{Dom}(\Sigma) \land \lambda(\bar{x}) = 1 \quad
\end{equation}

Any deviation $\Delta = \bar{x} \setminus \Sigma$ where $\Delta \neq \emptyset$ results in a set-theoretic mismatch, triggering a system exception. The complexity of integration grows linearly with the rigidity of $\Sigma$, creating what we term "Coupling Hysteresis"—the resistance of the system to conform to new input shapes without refactoring.

\subsection{The Liquid Paradigm: Intentional Probabilistic Negotiation}

A Liquid Interface is formally defined not as a stored object, but as a time-bounded generative function. Let $\Phi$ denote the \textit{Intention} (the semantic goal), $\mathcal{C}_t$ the temporal Context, and $\mathcal{G}$ the Governance constraints.

We posit that a Liquid Interface $I_{liquid}$ is the output of a negotiation function $\mathcal{N}$ executed at time $t_0$:

\begin{equation}
    I_{liquid}(t) = \mathcal{N}(A_{emitter}, A_{receiver}, \Phi, \mathcal{C}_t) \quad \text{subject to} \quad \mathcal{G}
\end{equation}

Unlike $I_{solid}$, which demands structural equality, $I_{liquid}$ seeks **Semantic Equivalence**. The function $\mathcal{N}$ operates as an optimization process that minimizes the \textit{Semantic Entropy} $H(S)$ between the agents' horizons:

\begin{equation}
    \min_{\pi} H(S) = -\sum_k p(m_k \mid \Phi, C_t) \log p(m_k \mid \Phi, C_t) \quad
\end{equation}

Where $\pi$ is the generated protocol (the temporary interface) and $p(m_k)$ is the probability that message $m_k$ correctly fulfills intention $\Phi$.

\subsection{Temporal Ephemerality and Debt Cancellation}

A defining characteristic of the Liquid Interface is its lifecycle. While $I_{\text{solid}}$ persists indefinitely ($\Delta t \to \infty$), $I_{\text{liquid}}$ is strictly ephemeral. We define the existence of the interface as a function of the interaction window $W$:

\begin{equation}
    \exists I_{\text{liquid}}(t) \iff t_{\text{start}} \leq t \leq t_{\text{ack}} \quad
\end{equation}

At $t > t_{\text{ack}}$ (post-execution), the interface structure $\pi$ dissolves. Mathematically, this implies that the Residual Coupling ($R_c$) between Agent A and Agent B becomes zero:

\begin{equation}
    \forall t > t_{\text{ack}}, \quad R_c(A, B) = 0 \quad
\end{equation}

This formalizes the claim that Liquid Interfaces do not accumulate technical debt, as no static schema remains to be maintained or versioned. The "integration" is re-instantiated ab initio for every new intention, ensuring maximum plasticity.
\section{Protocol and System Architecture}

To instantiate the Liquid Interface paradigm, we define a protocol specification and a coordination architecture that concretely realizes intention-driven interaction, semantic negotiation, and ephemeral interface life cycles within distributed systems. The reference architecture presented here is illustrative rather than prescriptive. It demonstrates feasibility without constraining valid implementations of the protocol.

Rather than encoding fixed call semantics or static service contracts \cite{fielding2000}, the proposed architecture treats interaction as a temporally bounded coordination process among autonomous agents operating under uncertainty.
The protocol defines a small set of coordination invariants—intent-first interaction, negotiated agreement before execution, and mandatory dissolution—while leaving matching, authorization policies, and conflict resolution strategies as replaceable policy choices.

Consequently, LIP is not intended to replace low-latency deterministic RPC or real-time control protocols, but to support coordination under semantic uncertainty in open agent environments.

\subsection{Liquid Interface Protocol (LIP)}

The Liquid Interface Protocol (LIP) defines a coordination model through which interfaces emerge, stabilize, and dissolve as a consequence of goal-directed interaction among autonomous agents. In contrast to conventional request--response protocols and message-oriented agent standards, LIP treats interaction as a negotiated, intention-driven process rather than as the invocation of predefined operations.

At its core, LIP governs how agents articulate intentions, expose capabilities, negotiate terms of collaboration, and coordinate execution under semantic uncertainty. The protocol privileges semantic alignment over syntactic compliance, allowing interaction to proceed without pre-established schemas, fixed endpoints, or globally shared ontologies.

LIP is realized over persistent bidirectional communication channels and enforces a constrained interaction structure that ensures progress while preserving interpretive flexibility. Execution is permitted only after sufficient semantic agreement has been  established, and all coordination artifacts are explicitly dissolved upon task completion, thereby enforcing ephemerality as a protocol-level invariant. This approach departs from traditional interface-centric integration models by treating coordination itself as a first-class, transient construct \cite{fielding2000,FIPA2002}.

\subsubsection{Message Semantics, Interaction Lifecycle, and Ephemerality}

All protocol messages in LIP are represented as structured JSON documents and are categorized by a \texttt{message\_type} field that determines their semantic role within an interaction. Message types are not merely communicative primitives, but performative acts that collectively define the lifecycle of a Liquid Interface.

The core message types are:

\begin{itemize}
    \item \textit{intent}: articulates a high-level objective and instantiates a new interaction context, giving rise to a Liquid Interface;
    \item \textit{offer}: declares the availability of a capability relevant to the expressed intention, potentially constituting a total or partial contribution toward its fulfillment;
    \item \textit{accept / reject}: signal negotiated agreement or refusal, contributing to the stabilization or revision of the interaction context;
    \item \textit{execute}: authorizes coordinated execution under the negotiated terms;
    \item \textit{complete}: signals termination of execution, independent of success or failure;
    \item \textit{dissolve}: explicitly invalidates the interaction context and triggers  mandatory cleanup of all coordination artifacts.
\end{itemize}

A Liquid Interface is defined as a temporally bounded coordination construct whose existence is contingent upon an active interaction context. Its lifecycle is not predetermined by static configuration, but emerges dynamically through the articulation of intent and subsequent semantic negotiation.

An interface is instantiated when an agent expresses an intention and at least one counterparty acknowledges semantic relevance. Stabilization does not imply commitment to a single executing agent; agreement may bind multiple partial offers into a composite execution structure whose coordinated application satisfies the declared intention.Execution is explicitly conditioned on this stabilization and may be aborted or renegotiated in response to failure, constraint violation, or withdrawal of consent.

Upon completion or abandonment of execution, a \textit{dissolve} message enforces ephemerality by invalidating all coordination artifacts, including negotiated agreements, authorization scopes, and execution bindings. No residual interface structures persist beyond this point.

By defining lifecycle semantics through message meaning rather than explicit control states, LIP treats interfaces as transient coordination events rather than persistent integration artifacts. This explicit instantiation and mandatory dissolution prevent the accumulation of long-lived coupling and preserve the
protocol’s commitment to minimal residual state.

\begin{figure}[!htp]
  \centering
  \begin{tikzpicture}[
    layer/.style={
      draw,
      thick,
      rounded corners,
      align=center,
      minimum width=11cm,
      minimum height=1.7cm,
      font=\sffamily
    }
  ]

  \node[layer] (agents) at (0,5.6)
    {Application-Level Agents \& Business Logic\\
     \footnotesize Autonomous agents exposing capabilities and intentions};

  \node[layer] (lip) at (0,3.6)
    {Liquid Interface Protocol (LIP) Coordination Layer\\
     \footnotesize Intent, discovery, negotiation, claim-based authorization, dissolution};

  \node[layer] (transport) at (0,1.6)
    {Transport Layer\\
     \footnotesize WebSockets / QUIC / gRPC / MCP; bidirectional message delivery};

  \node[layer] (crypto) at (0,-0.4)
    {Cryptographic Identity \& Claims\\
     \footnotesize Enrollment, challenge--response, claims, signatures};

  \node[layer] (infra) at (0,-2.4)
    {Infrastructure Layer\\
     \footnotesize Compute cluster, mesh, federation / WAN substrate};

  \draw[->, thick] (infra) -- (crypto);
  \draw[->, thick] (crypto) -- (transport);
  \draw[->, thick] (transport) -- (lip);
  \draw[->, thick] (lip) -- (agents);

  \end{tikzpicture}
  \caption{Reference stack underlying the Liquid Interface Protocol (LIP), illustrating the separation
  between semantic coordination, transport delivery, cryptographic identity, and infrastructure substrate.}
  \label{fig:lip_reference_stack}
\end{figure}
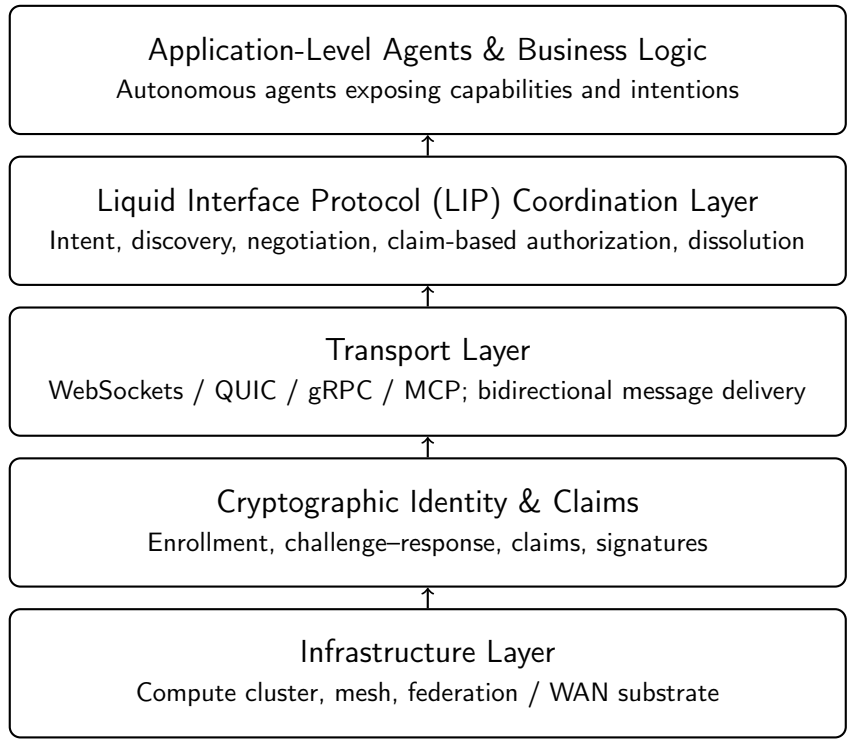

\begin{figure}[!htp]
  \centering
  \includegraphics[width=\linewidth]{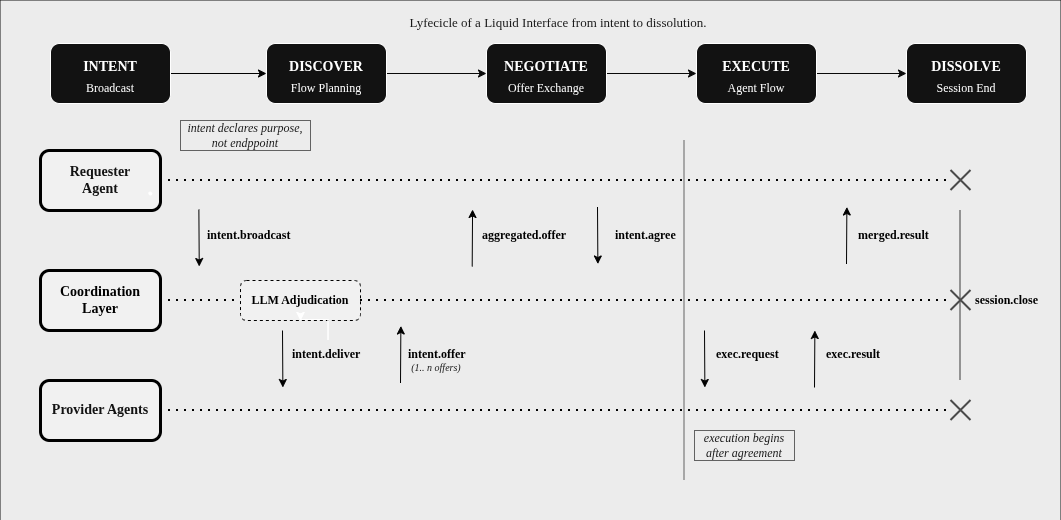}
  \caption{Macro-architecture of the Liquid Interface Protocol (LIP).}
  \label{fig:lip_intent_lifecycle}
\end{figure}

Beyond execution semantics, LIP presupposes a layered substrate that supports semantic coordination, transport delivery, cryptographic identity, and physical deployment. Figure~\ref{fig:lip_reference_stack} illustrates this reference stack, while Figure~\ref{fig:lip_intent_lifecycle} summarizes the lifecycle of a Liquid Interface as instantiated through the protocol’s performative message types. The interface emerges upon articulation of intent, stabilizes through negotiation, executes under the negotiated terms, and is mandatorily dissolved at the end of the session, ensuring no persistent coordination artifacts remain
beyond the interaction context.

\FloatBarrier
\subsubsection{Semantic Capability Matching via LLM Adjudication}

Semantic capability matching in the Liquid Interface Protocol is formulated as a \emph{decision problem under uncertainty}, rather than as a purely retrieval-based similarity task. Given an expressed intention $\Phi$, a contextual state $C_t$, and a set of declared agent capabilities $\{c_i\}$, the coordination layer must determine which agents are semantically and operationally compatible with the intended objective.

In the current instantiation, this decision is performed through an \emph{LLM-based semantic adjudicator}. The adjudicator is provided with a structured representation of the intention, contextual constraints, and capability descriptions, and is tasked with producing an ordered set of candidate agents along with an explicit acceptability judgment for each candidate. The adjudication process evaluates not only topical relevance, but also constraint satisfiability, implicit dependencies, and contextual alignment.

Formally, the adjudicator implements a probabilistic decision function:

\begin{equation}
f_{\text{LLM}} : (\Phi, C_t, \{c_i\}) \rightarrow \{(c_i, s_i, r_i)\}
\end{equation}

where $s_i \in [0,1]$ denotes the estimated suitability score of capability $c_i$ for fulfilling intention $\Phi$, and $r_i$ is a structured justification trace describing the rationale for the decision.

Candidates whose suitability scores exceed a configurable threshold are admitted into the negotiation phase, while others are excluded. This mechanism enables the coordination layer to reason over unstructured and partially specified capabilities, supporting interaction between agents with no prior coordination history.

By framing matching as semantic adjudication rather than lexical similarity, the protocol accommodates ambiguity, conflicting constraints, and underspecified objectives—properties that are intrinsic to real-world coordination problems. Moreover, the adjudicator abstraction remains orthogonal to the protocol itself, allowing alternative instantiations (e.g., embedding-based retrieval or hybrid pipelines) without altering the interaction semantics defined by LIP.

Recent work has demonstrated that large language models can act as semantic evaluators and decision layers in tool-augmented and agentic systems, supporting reasoning over constraints and action feasibility \cite{schick2024toolformer,yao2023tree,xi2023agents}.

In addition to semantic adjudication, LIP supports outcome-informed ranking as a secondary signal for candidate selection. Historical interaction outcomes are used to derive soft performance priors associated with individual agents, reflecting factors such as successful task completion, response quality, and execution latency.

These priors do not override semantic compatibility, but bias the adjudication process when multiple candidates exhibit comparable suitability. Formally, the adjudicator may incorporate an outcome-based prior $\rho_i$ for each capability $c_i$, learned from past interactions, such that:

\begin{equation}
s_i' = \alpha \cdot s_i + (1 - \alpha) \cdot \rho_i
\end{equation}

where $s_i$ denotes the semantic suitability score produced by the adjudicator, $\rho_i$ represents the normalized historical performance prior, and $\alpha \in [0,1]$ controls the relative influence of semantic alignment versus empirical outcome.

Outcome priors may be derived from multiple signals, including execution success, response consistency, explanation coherence, and interaction latency, and are updated incrementally over time. By design, these priors remain advisory rather than deterministic, ensuring that agents are not permanently penalized for transient
failures and that novel or infrequently used agents remain discoverable.

This mechanism allows LIP to balance semantic intent alignment with empirical performance, enabling adaptive coordination without introducing static trust hierarchies or persistent coupling.

\subsubsection{Failure Handling and Renegotiation}

Failure is treated in LIP as an expected and informative outcome of coordination, rather than as an exceptional condition. Given the open and semantically uncertain nature of agent interactions, failures may arise from unavailable capabilities, unsatisfied constraints, insufficient authorization scopes, or execution-time
conditions beyond the control of any single participant.

Rather than terminating the interaction upon failure, LIP allows failure signals to trigger renegotiation. When execution cannot proceed or complete under the negotiated terms, agents may revise constraints, withdraw offers, or propose alternative capabilities without invalidating the interaction context. In this sense, failure functions as a coordination signal that prompts semantic realignment rather than protocol-level abort.

Failure handling preserves the protocol’s core invariants: execution is never permitted without sufficient agreement, and no partial or ambiguous outcomes are committed as stable interface artifacts. If renegotiation does not yield a viable configuration, the interaction is explicitly dissolved, ensuring that no residual
state persists beyond the failed attempt.

By elevating failure to a first-class semantic event, LIP supports resilient coordination under uncertainty while avoiding exception-driven control flow and long-lived error states common in request--response integration models.

\subsubsection{Security and Authorization}

Security in the Liquid Interface Protocol is defined at the level of agent interaction,
independent of any specific coordination infrastructure. The protocol adopts a layered
security model suitable for open multi-agent environments, where participants may have
no prior trust relationship and interaction contexts are inherently ephemeral.

Agent authentication follows a two-phase process: \emph{enrollment} and
\emph{authentication}. During enrollment, an agent requests registration by submitting
a public key that uniquely represents its cryptographic identity. Upon acceptance,
this public key is recorded as the persistent identifier of the agent within the protocol
context.

Subsequent authentication is performed through a challenge--response mechanism.
When an agent initiates a new interaction, the coordinating party issues a cryptographic
challenge that must be signed using the agent’s private key. Successful verification
establishes agent authenticity without disclosure of secret material, ensuring
resistance to replay and impersonation attacks.

Beyond identity verification, the protocol enforces message-level integrity by requiring
cryptographic signatures on semantically binding messages (e.g., \texttt{accept},
\texttt{execute}, \texttt{complete}), thereby providing non-repudiation and preserving
the integrity of negotiated agreements.

Authorization in LIP is modeled as a \emph{claim-based, policy-governed process} rather
than as a static access-control list. Agents may declare (\emph{claim}) authorization
scopes representing the classes of actions or capabilities they intend to exercise.
These claims are evaluated against organizational-defined authorization policies before
being approved for participation in the negotiation or execution phases.

Authorization scopes are limited to the interaction context and may be constrained by
organizational boundaries, agent roles, trust levels, or contextual conditions. Approved
scopes grant permission to engage in specific categories of coordination actions but do
not constitute permanent privileges. All authorization grants are explicitly bound to
the lifetime of the interaction and are revoked upon dissolution.

This authorization model aligns with previous work on capability and claim-based
security in distributed systems, while extending these concepts to semantically
negotiated and ephemeral interaction contexts \cite{lampson1992access}.
\section{The Logic of Ephemeral Coordination}
\label{sec:logic}

To operationalize the transition from static interfaces to liquid coordination, we
define the \emph{Liquid Interface Protocol} (LIP) not as a software artifact, but as a set of architectural invariants and logical constraints governing agent interaction under semantic uncertainty. In this respect, LIP plays a role analogous to the constraint-based formulation of REST, which defined the architectural properties of the Web without prescribing specific implementations \cite{fielding2000}.

Rather than validating interactions through syntactic conformance alone, LIP
characterizes coordination as a bounded, goal-directed process whose validity
depends on semantic alignment, negotiated agreement, and enforced ephemerality.
This section formalizes the core coordination invariants, the negotiation function,
and the boundary conditions that preserve system safety when semantic negotiation
fails.

\subsection{Coordination Invariants}

A system may be classified as implementing a Liquid Interface if and only if it
satisfies the following coordination invariants. These constraints are intentionally
\emph{negative}: they restrict classes of behavior that would otherwise reintroduce
static coupling or long-lived integration debt.

\subsubsection{Invariant I: Intentional Primacy}

In conventional interface-driven architectures, interaction validity is determined by syntactic compliance: $V(x) \iff x \in \Sigma$. In contrast, LIP defines validity relative to the declared intention $\Phi$. An interaction step is considered valid only if it advances semantic alignment with respect to that intention.

Formally, let $H(S)$ denote the entropy of the shared semantic state between two agents. An interaction step at time $t$ is valid if and only if:

\begin{equation}
    \Delta H(S) = H(S)_t - H(S)_{t+1} > 0 \quad
\end{equation}

where $\Delta H(S) = H(S)_t - H(S)_{t+1}$, so a positive value indicates reduced uncertainty.

That is, the interaction must reduce semantic uncertainty rather than merely satisfy
structural constraints. Messages that are syntactically valid but fail to improve
semantic alignment are treated as coordination failures, triggering clarification or
renegotiation rather than execution. This reframes error handling as a semantic
breakdown rather than a protocol exception \cite{Gadamer1960}.

\subsubsection{Invariant II: Mandatory Temporal Dissolution}

A defining property of liquid coordination is the prohibition of persistent interface structures. Any negotiated interface $I_{liquid}$ is strictly bound to the temporal window of the interaction:

\begin{equation}
    \exists I_{\text{liquid}}(t) \iff t_{\text{start}} \leq t \leq t_{\text{ack}} \quad
\end{equation}

Upon completion or abandonment of execution at $t_{ack}$, all negotiated schemas,
authorization scopes, and execution bindings are invalidated. Persistence of these
artifacts beyond the interaction window constitutes a violation of the protocol.

This invariant prevents the gradual accumulation of implicit contracts and stale
assumptions, enforcing ephemerality as a first-class coordination property rather
than an implementation detail \cite{Lehman1980}.

\subsubsection{Invariant III: Bounded Residual Coupling}

To preserve loose coupling under repeated interaction, LIP constrains the residual dependency between agents across coordination episodes. Let $R_c(A, B)$ denote the residual coupling induced by prior interactions between agents $A$ and $B$. LIP requires that:

\begin{equation}
    \forall t > t_{\text{ack}}, \quad R_c(A, B) = 0 \quad
\end{equation}

This invariant does not forbid learning or adaptation, but prohibits the reuse of
previously negotiated interface structures as binding assumptions. Each interaction must be renegotiated, ensuring robustness against concept drift, capability evolution, and contextual change \cite{Latour2005}.

\subsubsection{Invariant IV: Composable Intent Resolution}

Liquid coordination does not require that a single agent satisfies a declared intention. An intention $\Phi$ can be resolved through the composition of multiple partial capabilities, potentially contributed by distinct agents, provided that their ordered execution advances semantic alignment with respect to $\Phi$.

Formally, let ${\Phi_1, \ldots, \Phi_n}$ denote a set of sub-intentions such that their sequential or partially ordered execution reduces semantic entropy relative to the original intention:

\begin{equation}
    \sum_{i=1}^{n} \Delta H(S_i) > 0 \implies \Phi \text{ is resolvable} \quad
\end{equation}

Agreement under LIP is reached over the composite execution structure as a whole, rather than over individual steps in isolation. Partial acceptance of constituent capabilities does not authorize execution; execution is permitted only after atomic agreement on the full composition.

This invariant guaranties that multi-agent coordination emerges from negotiated capability composition rather than centralized orchestration, while preserving the ephemeral and decoupled nature of liquid interfaces.

\subsection{The Negotiation Function}

The core operational mechanism of LIP is the negotiation function $\mathcal{N}$, which synthesizes a temporary coordination structure at runtime. Rather than retrieving predefined interfaces, $\mathcal{N}$ generates a provisional protocol $\pi$ that minimizes semantic uncertainty subject to contextual and governance constraints.

\begin{equation}
    I_{\text{liquid}} = \mathcal{N}(\Phi, C, \mathcal{G}) = \arg \min_{\pi} \left\{ \sum_k -p(m_k \mid \Phi) \log p(m_k \mid \Phi) : \pi \models \mathcal{G} \right\} \quad
\end{equation}

Here, $p(m_k \mid \Phi)$ denotes the estimated probability that message structure $m_k$ advances satisfaction of the declared intention. Negotiation proceeds iteratively until semantic confidence exceeds a predefined safety threshold $\tau$, at which point execution may commence.

\subsection{Boundaries of Liquidity and Deterministic Fallbacks}

A complete theory of liquid coordination must explicitly account for failure modes. We define the \textit{boundary of liquidity} as the point at which semantic negotiation fails to converge within acceptable temporal or uncertainty bounds.

\subsubsection{Entropy Threshold}

Let $H(S)$ denote the semantic entropy of the shared interpretive state and let
$t_{\text{start}}$ be the initial negotiation time. The initial entropy is defined as:
\begin{equation}
    H_0 = H(S)\big|_{t = t_{\text{start}}}.
\end{equation}

To ensure that semantic negotiation converges within acceptable bounds, we define a tolerance parameter $\tau \in (0, 1)$ representing the minimum required fractional reduction in entropy for negotiation to succeed. The maximum admissible entropy is thus:
\begin{equation}
    H_{\max} = (1 - \tau) H_0.
\end{equation}

Negotiation proceeds iteratively for at most $n_{\max}$ steps. The liquid state is
considered unstable if no iteration achieves entropy below the admissible threshold:
\begin{equation}
    \forall n \le n_{\max}, \quad H(S)_n > H_{\max} \;\;\implies\;\; \textsc{TriggerFallback}.
\end{equation}

This condition identifies a formal boundary at which interpretive flexibility becomes detrimental to liveness. For example, setting $\tau = 0.5$ requires negotiation to reduce entropy by at least 50\% of its initial value. Domain-specific safety policies may instead choose $\tau$ based on minimum semantic confidence required for the application domain.

\subsubsection{Fallback Modes}

Upon crossing the boundary of liquidity, LIP mandates deterministic reversal. Two fallback modes are defined:

\begin{enumerate}
    \item \textbf{Recursive Simplification}: The protocol retries negotiation with a simplified intention $\Phi' \subset \Phi$ or an expanded context $C'$, analogous to human rephrasing in cooperative dialogue.

    \item \textbf{Solidification}: The protocol abandons semantic negotiation and anchors interaction to a minimal, pre-agreed core ontology $\Sigma_{\text{core}}$, ensuring deterministic execution of critical actions.
\end{enumerate}

By explicitly defining these boundaries, LIP treats liquidity as a bounded and governed coordination regime rather than an unconstrained source of nondeterminism.
\section{Governance Implications of Liquid Interfaces}

The introduction of Liquid Interfaces has direct implications for how governance, security, and accountability are expressed in distributed systems. By elevating intent, negotiation, and semantic agreement to first-class protocol constructs, LIP enables governance mechanisms that operate at the level of meaning rather than at the level of static interfaces or predefined operations.

This section discusses the governance properties implied by the protocol, focusing on intention-based authorization, semantic auditability, and threat mitigation principles inherent to liquid coordination.

\subsection{Intention-Based Governance}

Traditional access-control models such as Role-Based Access Control (RBAC) and Attribute-Based Access Control (ABAC) authorize discrete operations against predefined resources. In contrast, Liquid Interfaces enable an
\emph{Intention-Based Access Control} (IBAC) model, in which governance decisions are evaluated against the expressed intent, its contextual constraints, and organizational policies.

Under IBAC, authorization is decoupled from concrete execution paths. Policies are applied prior to negotiation and execution, allowing organizations to govern \emph{what is being attempted} rather than \emph{how it is technically realized}. This shift enables consistent enforcement of business rules and compliance constraints even as execution plans, participating agents, or capabilities vary dynamically.

By aligning authorization with intent rather than endpoints, IBAC supports adaptive governance in environments characterized by semantic uncertainty and evolving coordination patterns.

This perspective builds on prior work that treats coordination as distinct from
computation, extending it into the domain of governance and access control under
semantic uncertainty \cite{papadopoulos1998coordination}.

\subsection{Semantic Auditability}

Because LIP treats intention articulation, negotiation, and execution as explicit protocol events, it enables a form of auditability that extends beyond traditional request--response logging. Instead of recording only syntactic operations, liquid interactions preserve the semantic context and rationale underlying coordination decisions.

Semantic auditability allows post-hoc reconstruction of \emph{why} a particular coordination outcome was reached, including the evaluated intent, negotiated terms, applied policies, and observed outcomes. This property is critical for accountability, regulatory compliance, and incident analysis in autonomous and semi-autonomous systems.

By capturing reasoning context rather than isolated actions, LIP provides a foundation for explainable governance in agent-mediated environments.

This form of auditability aligns with interpretive views of action and decision-making, in which meaning and context are inseparable from execution \cite{Winograd1986}.

\subsection{Threat Model Considerations}

The open and dynamic nature of liquid coordination introduces threat vectors that differ from those of traditional interface-centric systems. These include malicious capability claims, adversarial intent formulation, replay of negotiation messages, and attempts to exploit partial semantic agreement.

LIP addresses these risks through a combination of protocol-level principles: cryptographic identity binding, semantic validation of intents and capabilities, message-level integrity, and explicit lifecycle enforcement. Importantly, failure and renegotiation are treated as first-class outcomes, reducing reliance on exception-driven control flow that may otherwise amplify attack impact.

Mandatory dissolution of interaction contexts further limits the persistence of compromised coordination artifacts, constraining the blast radius of adversarial behavior. As a result, governance in LIP emphasizes containment, revocability, and semantic validation over static perimeter defenses.

These mitigation principles are consistent with foundational security design principles such as least privilege, fail-safe defaults, and containment \cite{Saltzer1975}.

\subsection{Governance as a Consequence of Coordination}

Taken together, these properties position governance in Liquid Interfaces not as an external control layer, but as an emergent consequence of semantically mediated coordination. Authorization, auditability, and risk mitigation arise from the same protocol mechanisms that enable negotiation and execution, preserving coherence
between governance objectives and operational behavior.

This integration allows organizations to govern autonomous agent interactions without reintroducing rigid interface contracts, maintaining alignment between liquid coordination and institutional accountability.

This view is consistent with sociotechnical perspectives in which governance and stability emerge from networks of interaction rather than from centralized control structures \cite{Latour2005}.
\section{Limitations and Open Challenges}

While Liquid Interfaces address fundamental limitations of static integration paradigms, the proposed model deliberately introduces a different set of trade-offs. The constraints discussed in this section are not incidental shortcomings of the protocol, but intrinsic boundaries arising from the choice to treat coordination as a semantic, intention-driven process rather than as a deterministic interface invocation mechanism. As such, they delineate the regimes in which liquid coordination is appropriate, as well as the open challenges that emerge from this reframing.

\subsection{Paradigm Boundaries}

Liquid Interfaces prioritize semantic flexibility, adaptive negotiation, and contextual interpretation over deterministic execution guarantees. Consequently, the paradigm is not suitable for domains that require strict real-time constraints, bounded worst-case latency, or hard safety guarantees enforced through static verification. Coordination mechanisms grounded in semantic negotiation inherently introduce deliberative overhead that cannot be eliminated without reverting to rigid interface contracts.

Moreover, Liquid Interfaces are designed for environments characterized by semantic uncertainty, evolving capabilities, and open-ended interaction. In domains governed by stable ontologies, well-defined schemas, and long-lived contracts, traditional interface-centric architectures may offer superior efficiency and predictability. In such settings, the overhead of dynamic negotiation may outweigh its benefits.

Finally, by shifting governance from static artifacts to contextual evaluation, the protocol may conflict with regulatory or organizational environments that mandate immutable contracts or pre-certified execution paths. Reconciling ephemeral coordination with such compliance requirements remains an open challenge and may require hybrid architectures that combine liquid negotiation with selectively solidified execution cores.

\subsection{Coordination Trade-offs}

The reference architecture discussed in this work adopts a logically centralized coordination substrate to simplify semantic mediation and negotiation ordering. While centralization is not a requirement of the Liquid Interface Protocol itself, fully decentralized or federated realizations introduce additional complexity related to semantic consistency, policy enforcement, and negotiation convergence.

In particular, decentralized coordination raises challenges in maintaining coherent semantic adjudication across heterogeneous agents, especially when contextual information is unevenly distributed. Designing coordination substrates that preserve the protocol’s semantic invariants without reintroducing rigid coupling or centralized control remains an open research direction.

\subsection{Trust and Capability Verification}

Liquid Interfaces rely on agents declaring capabilities and constraints as part of the negotiation process. While cryptographic identity binding ensures agent authenticity, it does not guarantee the veracity or completeness of declared capabilities. As a result, liquid coordination introduces a tension between openness and trust.

Mechanisms for capability verification, reputation modeling, and adversarial resilience must balance robustness against the protocol’s commitment to semantic flexibility and low coupling. Overly strict verification risks excluding novel or infrequently interacting agents, while overly permissive policies may enable strategic misrepresentation or exploitation of partial semantic agreement.

Designing verification and trust mechanisms that remain advisory rather than determinative—thereby avoiding the reintroduction of static trust hierarchies—remains an unresolved challenge.

\subsection{Dynamic Context and Renegotiation}

The protocol treats failure and renegotiation as first-class coordination outcomes. However, dynamic changes in context during execution—such as shifting constraints, partial fulfillment of intentions, or external environmental changes—raise questions about when renegotiation should be triggered and how existing agreements should be revised.

Establishing principled criteria for renegotiation without introducing excessive coordination overhead or oscillatory negotiation cycles is an important area for future work. These challenges are inherent to any coordination regime that operates under semantic uncertainty and cannot be fully resolved through static protocol design.

\subsection{Scope of Applicability}

Liquid Interfaces are not intended as a universal replacement for conventional integration mechanisms. Rather, they define a coordination regime optimized for interaction under semantic uncertainty among heterogeneous autonomous agents.

Identifying hybrid architectures in which liquid coordination coexists with static interfaces—and determining appropriate boundaries between dynamic negotiation and deterministic execution—remains an open research direction. Such hybrids are likely to be essential for practical adoption, enabling systems to selectively exploit liquidity where adaptability is required while retaining solid interfaces where stability and efficiency dominate.

\section{Conclusion}

This work introduced the concept of \emph{Liquid Interfaces} as a fundamental
reframing of system integration in the presence of autonomous agents and semantic
uncertainty. Rather than treating interfaces as static contracts between predefined
endpoints, we proposed an interaction model in which interfaces emerge dynamically
as transient coordination events, negotiated at runtime and dissolved upon task
completion.

Grounded in sociological, philosophical, and computational foundations, the Liquid
Interface Protocol (LIP) formalizes this paradigm through intention-driven
interaction, semantic adjudication, negotiated agreement, and enforced ephemerality.
By elevating intent, negotiation, and reasoning to first-class protocol constructs,
LIP enables coordination among heterogeneous agents without requiring shared
ontologies, rigid schemas, or long-lived integration artifacts.

We further examined the governance implications of this shift, showing how
intention-based authorization, semantic auditability, and lifecycle-bound coordination
allow accountability and control to emerge from the same mechanisms that enable
adaptive interaction. In contrast to interface-centric security models, governance in
Liquid Interfaces operates at the level of meaning and context, preserving flexibility
without sacrificing oversight.

Taken together, Liquid Interfaces represent a move away from integration as
infrastructure and toward coordination as a semantic process. This perspective
challenges long-standing assumptions in software engineering, particularly the
equation of reliability with rigidity, and opens new directions for the design of
agent-mediated systems capable of operating under ambiguity, change, and incomplete
knowledge.

As autonomous agents become increasingly embedded in organizational and
sociotechnical environments, the ability to coordinate without static coupling will
become essential. Liquid Interfaces offer a principled foundation for this transition,
inviting further exploration into decentralized realizations, verification
mechanisms, and hybrid architectures that combine liquid coordination with
conventional systems.

\bibliographystyle{plainnat}
\bibliography{references}

@book{Bauman2000,
  author    = {Bauman, Zygmunt},
  title     = {Liquid Modernity},
  year      = {2000},
  publisher = {Polity Press},
  address   = {Cambridge, UK},
  isbn      = {978-0745624099}
}

@misc{Buterin2014,
  title        = {A Next-Generation Smart Contract and Decentralized Application Platform},
  author       = {Buterin, Vitalik},
  year         = {2014},
  howpublished = {Ethereum White Paper},
  url          = {https://ethereum.org/en/whitepaper/}
}

@article{BernersLee2001,
  title     = {The Semantic Web},
  author    = {Berners-Lee, Tim and Hendler, James and Lassila, Ora},
  journal   = {Scientific American},
  volume    = {284},
  number    = {5},
  pages     = {34--43},
  year      = {2001},
  publisher = {Scientific American, a division of Nature America, Inc.}
}

@article{birrell1984implementing,
  title     = {Implementing Remote Procedure Calls},
  author    = {Birrell, Andrew D. and Nelson, Bruce J.},
  journal   = {ACM Transactions on Computer Systems},
  volume    = {2},
  number    = {1},
  pages     = {39--59},
  year      = {1984}
}

@phdthesis{fielding2000,
  author = {Fielding, Roy Thomas},
  title  = {Architectural Styles and the Design of Network-Based Software Architectures},
  school = {University of California, Irvine},
  year   = {2000}
}

@misc{FIPA2002,
  title        = {FIPA ACL Message Structure Specification},
  author       = {{Foundation for Intelligent Physical Agents}},
  year         = {2002},
  howpublished = {\url{http://www.fipa.org/specs/fipa00061/}},
  note         = {Standard SC00061G}
}

@book{Gadamer1960,
  title     = {Truth and Method},
  author    = {Gadamer, Hans-Georg},
  year      = {1960},
  publisher = {Bloomsbury Academic},
  address   = {London, UK},
  note      = {Translation revised by J. Weinsheimer and D. G. Marshall (2004)},
  isbn      = {978-0826476975}
}

@inproceedings{hartig2018semantics,
  title     = {Semantics and Complexity of GraphQL},
  author    = {Hartig, Olaf and Pérez, Jorge},
  booktitle = {Proceedings of the 2018 World Wide Web Conference},
  pages     = {1155--1164},
  year      = {2018}
}

@article{kadavath2022language,
  title  = {Language Models (Mostly) Know What They Know},
  author = {Kadavath, S. and others},
  journal = {arXiv preprint arXiv:2207.05221},
  year   = {2022}
}

@article{Lehman1980,
  title     = {Programs, Life Cycles, and Laws of Software Evolution},
  author    = {Lehman, Manny M.},
  journal   = {Proceedings of the IEEE},
  volume    = {68},
  number    = {9},
  pages     = {1060--1076},
  year      = {1980},
  publisher = {IEEE},
  doi       = {10.1109/PROC.1980.11805}
}

@article{liu2023lost,
  title  = {Lost in the Middle: How Language Models Use Long Contexts},
  author = {Liu, Nelson and others},
  journal = {arXiv preprint arXiv:2307.03172},
  year   = {2023}
}

@book{Latour2005,
  title     = {Reassembling the Social: An Introduction to Actor-Network Theory},
  author    = {Latour, Bruno},
  year      = {2005},
  publisher = {Oxford University Press},
  address   = {Oxford, UK},
  isbn      = {978-0199256044}
}

@article{lampson1992access,
  author  = {Lampson, Butler and Abadi, Martín and Burrows, Michael and Wobber, Edward},
  title   = {Authentication in Distributed Systems: Theory and Practice},
  journal = {ACM Transactions on Computer Systems},
  year    = {1992}
}

@misc{MCP2024,
  title        = {Model Context Protocol (MCP): An Open Standard for Connecting AI Assistants to Systems},
  author       = {{Anthropic}},
  year         = {2024},
  howpublished = {\url{https://modelcontextprotocol.io}},
  note         = {Accessed: 2024-12-30}
}

@book{Meyer1992,
  title     = {Eiffel: The Language},
  author    = {Meyer, Bertrand},
  year      = {1992},
  publisher = {Prentice Hall},
  address   = {Hemel Hempstead},
  isbn      = {978-0132479257},
  note      = {Foundational work on Design by Contract}
}

@misc{mohsin2025fundamental,
  title         = {On the Fundamental Limits of LLMs at Scale},
  author        = {Mohsin, Muhammad Ahmed and others},
  year          = {2025},
  eprint        = {2511.12869},
  archivePrefix = {arXiv},
  primaryClass  = {cs.CL}
}

@incollection{papadopoulos1998coordination,
  author    = {George A. Papadopoulos and Farhad Arbab},
  title     = {Coordination Models and Languages},
  booktitle = {Advances in Computers},
  editor    = {M. V. Zelkowitz},
  volume    = {46},
  pages     = {329--400},
  publisher = {Academic Press},
  year      = {1998},
  address   = {New York, NY, USA},
  doi       = {10.1016/S0065-2458(08)60208-9}
}

@book{Posta2023,
  title     = {Istio in Action},
  author    = {Posta, Christian and Malfertheiner, Rinor},
  year      = {2023},
  publisher = {Manning Publications},
  isbn      = {978-1617295829}
}

@article{Saltzer1975,
  title   = {The Protection of Information in Computer Systems},
  author  = {Saltzer, Jerome H. and Schroeder, Michael D.},
  journal = {Proceedings of the IEEE},
  year    = {1975}
}

@article{Schick2024Toolformer,
  title     = {Toolformer: Language Models Can Teach Themselves to Use Tools},
  author    = {Schick, Timo and others},
  journal   = {Advances in Neural Information Processing Systems},
  volume    = {36},
  year      = {2024}
}

@article{sculley2015hidden,
  title   = {Hidden Technical Debt in Machine Learning Systems},
  author  = {Sculley, David and others},
  journal = {Advances in Neural Information Processing Systems},
  volume  = {28},
  year    = {2015}
}

@book{Winograd1986,
  title     = {Understanding Computers and Cognition: A New Foundation for Design},
  author    = {Winograd, Terry and Flores, Fernando},
  year      = {1986},
  publisher = {Addison-Wesley},
  address   = {Reading, MA},
  isbn      = {978-0201112979}
}

@article{xi2023agents,
  title  = {The Rise and Potential of Large Language Model Based Agents},
  author = {Xi, Zhiheng and others},
  journal = {arXiv preprint arXiv:2309.07864},
  year   = {2023}
}

@article{yao2023tree,
  title        = {Tree of Thoughts: Deliberate Problem Solving with Large Language Models},
  author       = {Shunyu Yao and Dian Yu and Jeffrey Zhao and Izhak Shafran and Thomas L. Griffiths and Yuan Cao and Karthik Narasimhan},
  journal      = {arXiv preprint},
  volume       = {arXiv:2305.10601},
  year         = {2023},
  url          = {https://arxiv.org/abs/2305.10601}
}

@misc{zhang2025selforganizing,
  title         = {Self-Organizing Agent Network for LLM-based Workflow Automation},
  author        = {Zhang, Ke and Liu, Wei and Chen, Jian},
  year          = {2025},
  eprint        = {2508.13732},
  archivePrefix = {arXiv},
  primaryClass  = {cs.AI}
}

\end{document}